\documentclass[11pt]{article}

\usepackage[final]{acl}

\usepackage{times}
\usepackage{latexsym}

\usepackage[T1]{fontenc}
\usepackage[utf8]{inputenc}

\usepackage{microtype}

\usepackage{inconsolata}

\usepackage{graphicx}
\usepackage{subcaption} % 子图
\usepackage{bm}
\usepackage{amsfonts}
\usepackage{amsmath, amssymb} % 公式
\usepackage[table]{xcolor}
\usepackage{booktabs} % 表格
\usepackage{multirow} % 表格
\usepackage{color, xspace} % 空格

\usepackage{nccmath}
\usepackage{enumerate}
\usepackage{enumitem}
\usepackage{algorithm}
\usepackage{newfloat}
\usepackage{listings}
\usepackage{caption}
\usepackage[export]{adjustbox}
\usepackage{mathtools}
\usepackage{tabularx}

\usepackage{algpseudocode}% http://ctan.org/pkg/algorithmicx

\usepackage{tcolorbox}

\tcbset{
  mypromptstyle/.style={
    colback=white,       % 背景白色
    colframe=gray!60,    % 边框颜色
    boxrule=0.4pt,       % 边框粗细
    arc=3pt,             % 圆角
    left=4pt,            % 左边距
    right=4pt,           % 右边距
    top=3pt,             % 上边距
    bottom=3pt,          % 下边距
    fontupper=\scriptsize
  }
}

\usepackage{xparse}
\newenvironment{prompt}
{
    \begin{tcolorbox}[
        mypromptstyle,
        before upper={\obeylines\obeyspaces\parindent=0pt} % 保留换行和空格，取消缩进
    ]
}
{
    \end{tcolorbox}
}

\newcommand{\name}{GSAR\xspace}

\title{\name: Goal-State-Anchor Rewards for Mobile GUI Agents with Self-Evolving Data Synthesis}

\author{
\begin{tabular}{c}
Long Zhang$^{1}$ \quad
Yuhan Chen$^{2}$ \quad
Chaoran Zhang$^{1}$ \quad
Wanxia Cao$^{2}$ \quad
Kun Huang$^{2}$ \\
Pengzhi Gao$^{2}$ \quad
Wei Liu$^{2}$ \quad
Jian Luan$^{2}$ \quad
Chenliang Li$^{1}$ \quad
Lixin Zou$^{1}$\thanks{\ \ Corresponding author.}
\end{tabular}
\\ \vspace{.5mm}
\begin{tabular}{c}
$^1$Wuhan University \quad
$^2$Xiaomi Inc.
\end{tabular}
\\ \vspace{.5mm}
\begin{tabular}{c}
\texttt{\{zlongooo, chaoranzhang, cllee, zoulixin\}@whu.edu.cn}\\
\texttt{\{yuhanchen240, caowanxia123, huangkun813\}@gmail.com}\\
\texttt{\{gaopengzhi, liuwei40, luanjian\}@xiaomi.com}
\end{tabular}
\vspace{2mm} \\
}
\begin{document}

\maketitle

\begin{abstract}
Vision-Language Models (VLMs) based GUI agents stand to benefit significantly from online reinforcement learning (RL). However, their training is bottlenecked by two fundamental issues: current data synthesis methods for GUI Agents rely on specific environments and struggle to generate diverse data, while existing evaluators either suffer from limited scalability or provide inaccurate and unreliable reward signals. 
To overcome these challenges, we introduce \name (Goal-State-Anchor Reward), a RL reward framework that supports scalable task generation and delivers reliable reward signals for stable and efficient policy optimization. 
Our approach features self-evolving data synthesis, which produces multiple environments through task execution and generates diverse tasks and goal states.
Complementing this, a state-anchor mechanism automatically annotates task-relevant UI elements in successful goal states as reference anchors. 
During RL training, these reference anchors provide accurate, scalable reward signals that substantially enhance efficiency. Extensive evaluations demonstrate that our framework achieves over 90\% accuracy on offline trajectory verification and performs closest to rule-based methods. Furthermore, agents trained using our reward framework exhibit strong performance on both AndroidWorld and our constructed benchmark, establishing a scalable approach for GUI agent training. Our code is publicly available at \url{https://github.com/lossooo/GSAR}.
\end{abstract}
\section{Introduction}

Graphical User Interface (GUI) agents driven by Vision-Language Models (VLMs)~\cite{Bai2025Qwen25VLTR,Hurst2024GPT4oSC} demonstrate human-like competence in understanding and operating mobile device environments~\cite{Wang2025UITARS2TR,Ye2025MobileAgentv3FA}. As a promising paradigm for automating interactions across mobile and desktop platforms, GUI agents transform raw screen observations into actionable decisions through visual comprehension and functional reasoning. These agents operate without extensive fine-tuning or additional pretraining, highlighting their strong potential for broad real-world applications.

\begin{figure}[t]
  \centering
  \includegraphics[width=\linewidth]{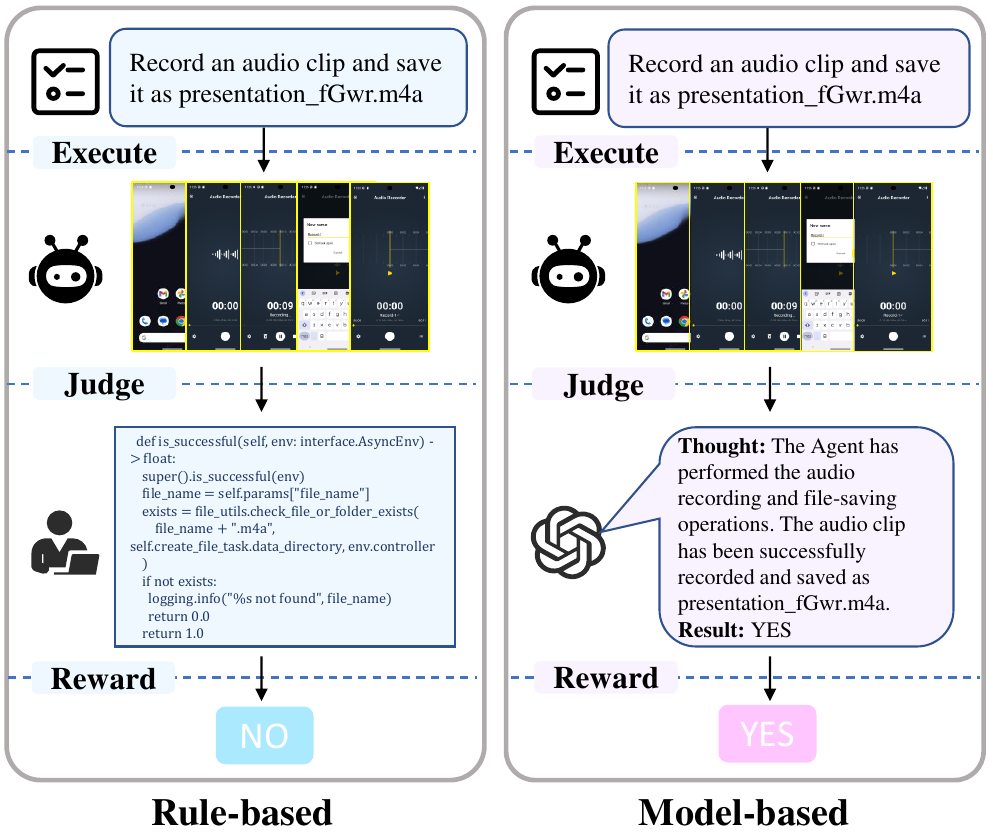}
  \caption{Rule-based and Model-based reward judge. Model-based judges may be inaccurate when context is limited or model capacity is insufficient.}
  \label{fig:reward_judge}
\end{figure}

Reinforcement Learning with Verifiable Rewards (RLVR)~\cite{Lambert2024TLU3P,Shao2024DeepSeekMathPT,Yu2025DAPOAO,Feng2025GroupinGroupPO}, building on its success in domains such as mathematics and code reasoning, has recently emerged as an effective paradigm for training GUI agents~\cite{Bai2024DigiRLTI,Xu2025MobileRLOA}. Scaling RLVR for GUI agents fundamentally relies on a unified training infrastructure: diverse task instantiation and strictly verified reward signals. While Large Language Models can trivially generate a vast array of task descriptions, grounding these open-ended instructions into executable training episodes requires configuring the corresponding initial environments. Currently, this environment setup and data collection process remains highly restricted by manually designed configurations~\cite{Sun2024OSGenesisAG,Lin2025GUIReWalkMD}. This labor-intensive dependency prevents the automatic generation of truly diverse environment-task pairs, which is the foundational prerequisite for large-scale policy learning.

Furthermore, evaluating agent performance across such an open-ended task space introduces an equally complex challenge. As shown in Figure~\ref{fig:reward_judge}, existing reward designs struggle to scale effectively. \textbf{Rule-based} methods require manually written execution functions to verify task completion~\cite{Rawles2024AndroidWorldAD,Xie2024OSWorldBM}; while precise, they are completely impractical to manually scale alongside automatically generated tasks. Conversely, \textbf{model-based} evaluators attempt to automate outcome assessment~\cite{Bai2024DigiRLTI,Wang2024DistRLAA}, but they frequently suffer from instability, hallucination, or insufficient coverage of edge cases. These flaws lead to noisy supervision signals that severely impair policy convergence and degrade agent performance. Together, the inability to automatically instantiate diverse initial environments and the lack of scalable, accurate reward supervision form a systemic bottleneck that constrains the true potential of RL for GUI agents.

To address this challenge, we introduce \name (Goal-State-Anchor Reward), an RL reward framework that tackles both bottlenecks by enabling scalable task and environment generation and providing reliable reward signals for stable policy optimization. First, we leverage self-evolving data synthesis to generate diverse tasks and trajectories, while progressively evolving the environments through interaction, resulting in task, initial environment, and trajectory triplets in a fully automated manner. Next, the state-anchor mechanism automatically annotates task-relevant UI elements in the final states of successful trajectories, producing task, initial environment, and goal-state-anchor reference triplets. During online training, these references are used to provide accurate and semantically grounded reward signals. Both online and offline experiments demonstrate that our approach, by explicitly anchoring task-relevant UI elements in successful goal states, delivers robust and scalable reward feedback for GUI agent training.

In summary, our work makes the following primary contributions:
\begin{itemize}[leftmargin=*]
\item We propose a self-evolving data synthesis mechanism that generates diverse environments and task trajectories while preserving initial environment snapshots, enabling reproducible and scalable reinforcement learning.
\item We introduce a scalable Goal-State-Anchor reward framework for GUI agents that overcomes the limitations of prior judges to deliver accurate, generalizable signals for online reinforcement learning.
\item We analyze the impact of reward accuracy on reinforcement learning and show that reward design critically affects agent performance.
\end{itemize}
\section{Related Works}

\subsection{Data Synthesis for Mobile GUI Agents}
Effective training of GUI agents requires diverse tasks that expose models to a wide range of interface elements and interaction patterns necessary for robust and generalizable performance. Following the introduction of sequential GUI data for mobile applications in Rico~\cite{Deka2017RicoAM}, research has increasingly focused on synthesizing GUI datasets. Several efforts~\cite{Rawles2023AndroidIT,Li2024OnTE,Lu2024GUIOA} generated data for supervised fine-tuning (SFT)~\cite{Ouyang2022TrainingLM} models, but the scalability of manually curated datasets is hindered by costly and inefficient data collection. More recently, automated data synthesis approaches have been proposed to address this limitation. These methods first explore application interfaces via random or heuristic traversal, and then construct tasks based on the explored states, either by reverse-engineering instructions~\cite{Sun2024OSGenesisAG} or generating tasks directly from screenshots~\cite{Xie2025GUIexplorerAE}. Large pretrained models (e.g., GPT-4o~\cite{Hurst2024GPT4oSC}) are then employed to execute these tasks and filter out incomplete or failed trajectories. However, these approaches still depend on manually preconfigured app environments and do not provide reproducible initial states required for stable reinforcement learning of GUI agents.

\begin{figure*}[t]
\centering
\includegraphics[width=\textwidth]{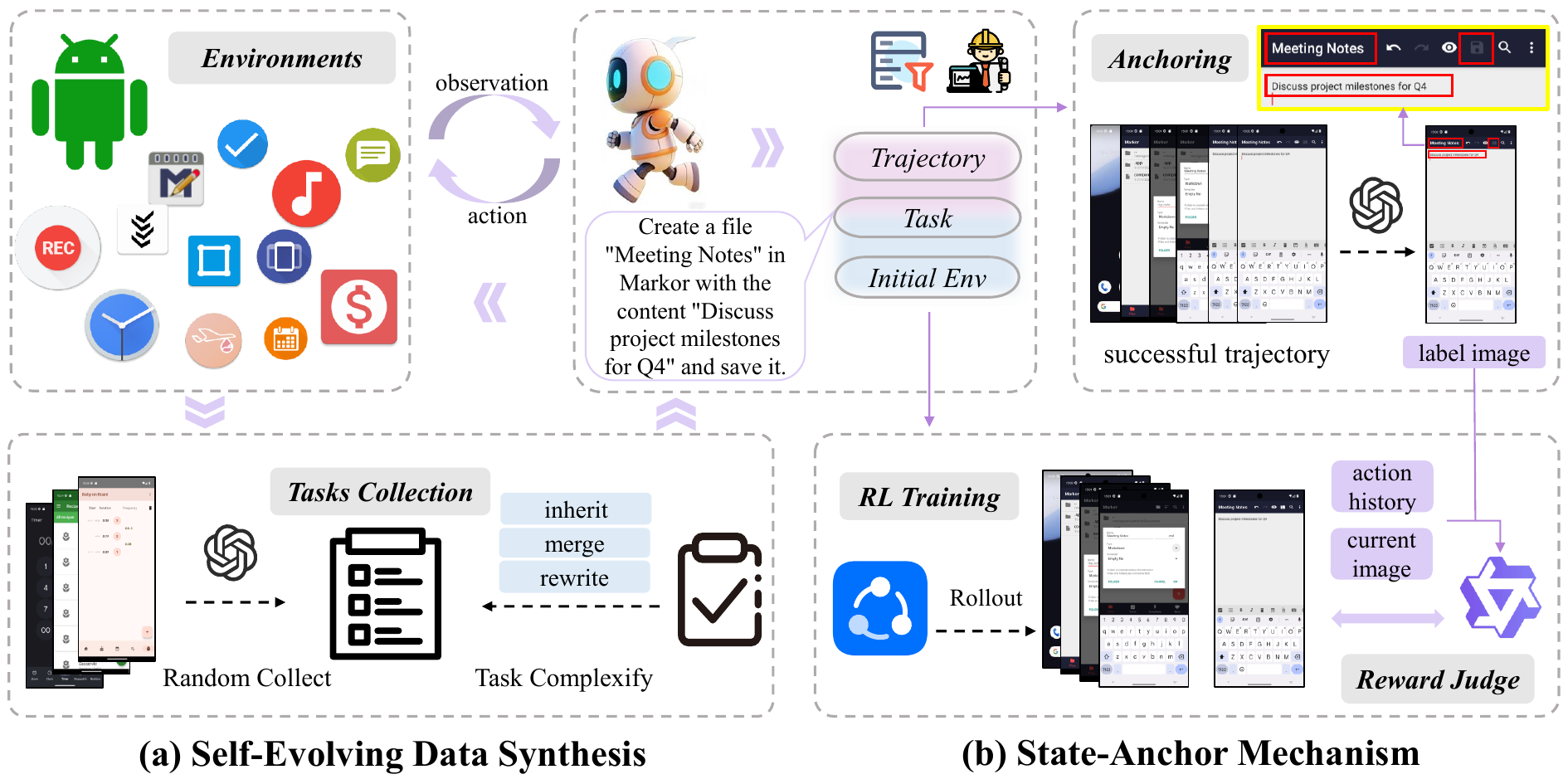}
\caption{Overview of \name. (a) self-evolving data synthesis interacts with mobile environments to modify environment states and automatically generate tasks and trajectories. (b) state-anchor mechanism leverages successful trajectories to identify and anchor task-related UI elements in the goal state, providing more accurate reward signals for training.}
\label{fig:method_overview}
\end{figure*}

\subsection{Reward System for GUI Agents}

Recent research has increasingly integrated reinforcement learning (RL) into the training of GUI agents~\cite{Zhang2025AgentCPMGUIBM,Chen2025STEPST}, with the goal of reducing supervision requirements and enhancing generalization and long-term decision-making ability. In RL, reward signals are crucial for guiding policy optimization and shaping learning dynamics~\cite{Gao2024OnDE,DeepSeekAI2025DeepSeekR1IR,Huang2025VisionR1IR}. Initial reward designs for GUI agents primarily focused on grounding tasks~\cite{Yuan2025EnhancingVG,Zhou2025GUIG1UR}, such as click position prediction~\cite{Lu2025UIR1EE} or intersection over union (IoU) measurement between predicted and target boxes~\cite{Liu2025InfiGUIR1AM}. Yet these methods are limited in scope and lack validation on more complex high-level tasks. To support training of high-level GUI agents, subsequent studies have developed rule-based outcome rewards~\cite{Rawles2024AndroidWorldAD,Luo2025GUIR1A} that provide reliable supervision through task-specific verification scripts. In addition, model-based reward functions have been explored for high-level tasks~\cite{Yang2025ZeroGUIAO,Shi2025MobileGUIRLAM}, where an external model evaluates successful demonstrations or outcomes, providing rewards with high scalability across tasks.

Building upon model-based methods, we propose a reward framework designed to provide more accurate and reliable reward signals for RL training while maintaining scalability.
\section{Method}
In this section, we introduce \name\ (Goal-State-Anchor Reward) framework. An overview of the \name\ pipeline is illustrated in Figure~\ref{fig:method_overview}. 
The framework operates through a seamless integration of two core phases. First, through \textbf{self-evolving data synthesis}, the framework progressively generates diverse tasks, execution trajectories, and their corresponding initial environments via interaction-driven evolution. Subsequently, the \textbf{state-anchor mechanism} leverages these trajectories for automatic reward annotation. By anchoring to goal states, this mechanism provides reliable and scalable reward supervision to close the learning loop.

\subsection{Self-Evolving Data Synthesis}
To synthesize data for training and to provide high-quality source data for reward annotation, we propose a self-evolving data synthesis framework that is both automatic and scalable. Unlike static or one-shot data generation pipelines, our approach explicitly models data construction as an evolving process. The key idea is to allow the task distribution, environment states, and trajectories to evolve together over time through continuous interaction with the GUI environment, closely mirroring how real-world applications are explored and used by humans. Task generation, execution, and trajectory collection are all driven by the model itself, enabling continuous expansion of the dataset without manual intervention. The self-evolving data synthesis framework is guided by the following principles:

\paragraph{Simulation of Real-World Usage.}
If an app has not undergone manual configuration or real user interaction, the information contained in a freshly installed app is typically minimal and simplistic. To enable data collection in richer and more diverse environments, we modify the app state by executing multiple tasks. Through repeated rounds of task execution and environment updates, the system progressively evolves the app states. This iterative process produces a wide variety of environments, which subsequently support the collection of more diverse tasks.

\paragraph{Automatic Task Generation.}
We first explore the GUI environment of each app through random interaction, producing an exploration trajectory:
\[
T = (s_0, a_0, s_1, a_1, \ldots, s_n),
\]
where $s_i$ denotes the environment state represented by the GUI screenshot, and $a_i$ represents the action taken at step $i$.

Given the explored observations along the trajectory, a Vision-Language Model $\mathcal{M}$ generates a set of potential user tasks based on the visual input:
\[
\mathcal{Q}_i = \mathcal{M}(s_i), \quad i \in \{0, \ldots, n\},
\]
where $\mathcal{Q}_i = \{q_i^1, q_i^2, \ldots, q_i^k\}$ denotes the set of generated tasks conditioned on observation $s_i$.

For each generated task, we preserve the corresponding environment \(e\) as the initial state, forming task–environment pairs:
\[
\mathcal{D}_{\text{task}} = \{(q, e) \mid q \in \mathcal{Q}_i \}.
\]
The resulting pairs are used to initialize reinforcement learning training episodes.

\paragraph{Trajectory Collection and Filtering.}
After task creation, we employ GPT-4o to execute these tasks within their associated environments in order to obtain trajectory data while simultaneously altering the app states. Due to the inherent limitations of the model, some generated tasks may not be executable under the current GUI context, and certain collected trajectories may contain mistakes or incomplete steps. To mitigate this issue, we adopt the filtering strategy $\mathcal{F}$ proposed in OS-Genesis~\cite{Sun2024OSGenesisAG} to remove task–trajectory mismatches. As a result, we obtain triplets consisting of task \(q\), initial environment \(e\), and successful trajectory \(\tau\):
\[
\mathcal{D}_{\text{traj}} = \{(q, e, \tau) \mid \mathcal{F}(q,\tau)\}.
\]

\paragraph{Progression from Simple to Complex.}
Although numerous tasks can be produced from individual states, we observe that the model struggles to generate tasks that are both sophisticated (e.g., involving multiple sub-goals or strict execution order) and diverse (e.g., similar tasks with varying parameters). To address this limitation, we apply a task complexification process to tasks that were successfully completed in the previous iteration.

Given a completed source triplet $(q_s, e_s, \tau_s)$, where $q_s$ denotes the source task instruction, $e_s$ is the initial environment, and $\tau_s$ represents the successful execution trajectory, we generate a new task $q_{\text{complexify}}$ through one of three strategies—inheritance, composition, or rewriting:
\[
q_{\text{complexify}} =
\left\{
\begin{aligned}
& \mathcal{M}_{\text{inherit}}(q_s, \tau_s), \quad e = e_T, \\
& \mathcal{M}_{\text{merge}}(q_s, \tau_s), \quad e = e_s, \\
& \mathcal{M}_{\text{rewrite}}(q_s, \tau_s), \quad e = e_s .
\end{aligned}
\right.
\]
Where $e_T$ denotes the initial environment of the current iteration. $\mathcal{M}_{\text{inherit}}$ generates a follow-up task starting from $e_T$, $\mathcal{M}_{\text{merge}}$ augments the source task by combining it with additional subtasks starting from the original environment, and $\mathcal{M}_{\text{rewrite}}$ produces a task variant by modifying specific parameters of the source task while preserving its overall structure. Through this process, the task distribution progressively evolves from simple tasks toward more complex and diverse workflows, generating increasingly challenging tasks across iterations. 

Further details and prompts for self-evolving data synthesis are provided in Appendix~\ref{appendix:self_evolving} and Appendix~\ref{appendix:prompts}.

\subsection{State-Anchor Reward}
While self-evolving data synthesis enables large-scale task and trajectory generation, effective online RL further requires accurate and scalable reward supervision. To this end, we introduce a state-anchor mechanism that supports automatic annotation with minimal human intervention.

\paragraph{Limitations in Action History Based Evaluation.}
In model-based reward evaluation, action history is often incorporated as contextual information to provide useful signals (e.g., the completion of intermediate steps), which helps the model better assess task progress. However, due to the limitations of current models, evaluators relying on historical context frequently suffer from false positives (FP). This issue is particularly pronounced in GUI scenarios, where visually similar states may correspond to semantically different task outcomes. Compared with rule-based approaches, model-based methods are less reliable because the model lacks explicit references of completed task states, making it difficult to determine whether a task has truly been accomplished. Inspired by this observation, we introduce goal-state references that explicitly represent the expected outcome of a task. Specifically, we annotate key elements within the goal state and use them as reference signals, providing the evaluator with a form of “ground-truth answer” that improves the accuracy of model-based reward estimation.

\paragraph{Goal-State Acquisition.}
We obtain goal states from the self-evolving data synthesis stage by extracting the final screenshot of successful trajectories. Formally, given a successful task–trajectory triplet $(q, e, \tau)$ where $\tau = (s_0, a_0, \ldots, s_T)$, the goal state is defined as the terminal observation:
\[
s_g = s_T .
\]

For certain difficult tasks, models may fail to complete them during trajectory collection, leading to their removal during filtering. However, such tasks may still be beneficial for reinforcement learning. Therefore, for a subset of these challenging tasks, the final goal states are obtained through manual execution in order to provide reliable completion references.

\paragraph{Automatic Reward Annotation via Goal-State Anchoring.}
After obtaining the goal state, we automatically identify UI elements relevant to the task objective and use them to construct a goal-state-anchor reference. These anchored elements provide a structured representation of the expected task outcome.

Specifically, given the goal state $s_g$ and its corresponding accessibility tree (a11y tree) $A_g$, we first overlay element indices from $A_g$ onto the screenshot and then use $\mathcal{M}$ to identify the indices of task-relevant elements:
\[
K_g = \mathcal{M}(\mathcal{I}(s_g, A_g), A_g),
\]
where $\mathcal{I}$ denotes the operation of overlaying element indices on the screenshot. The a11y tree provides semantic and spatial information for linking UI elements to their corresponding indices.

We then retrieve the bounding boxes of the selected elements from $A_g$ and use $\mathcal{A}$ to anchor their corresponding regions in the goal-state screenshot:
\[
G = \mathcal{A}\left(s_g, \{\operatorname{bbox}(A_g^k) \mid k \in K_g\}\right).
\]
The resulting $G$ serves as the goal-state-anchor reference for reliable task completion verification. Importantly, the entire process is human-free, enabling large-scale and continual reward annotation.

\paragraph{Goal-State-Anchor Reward for Training.}
Once the anchored goal state is obtained, it can serve as a reference answer to assist the model in evaluating task completion. In addition, inspired by~\cite{Lai2025AndroidGenBA}, we incorporate the action history as contextual information during evaluation. By combining the action history with the goal-state-anchor reference, the evaluator is provided with sufficient information to make more accurate judgments about task outcomes.
 
Formally, at time step $t$, the evaluator receives the current GUI state $s_t$, the action history $\tau_{0:t}$, and the goal-state-anchor label representation $G$, and outputs a binary reward indicating task completion:

\[
y_t = \mathcal{E}(s_t, \tau_{0:t}, G), \quad
r_t =
\begin{cases}
1, & \text{if } y_t = 1, \\
0, & \text{otherwise},
\end{cases}
\]

where 
\(
\mathcal{E}
\) denotes the evaluation model, 
\(y_t \in \{0,1\}\) indicates whether the task goal has been achieved, 
\(r_t\) is the reward at step \(t\).

During training, this reward mechanism can be applied at any step of task execution to determine whether the desired outcome has been achieved, enabling reliable reward supervision for RL.
\section{Experiments}
\begin{table*}[ht]
\centering
\resizebox{\linewidth}{!}{
\begin{tabular}{l@{\hspace{10pt}}cccccc@{\hspace{20pt}}c}
\toprule
\multirow{2}{*}{\textbf{Model}} & \multirow{2}{*}{\textbf{Data Sources}} 
& \multicolumn{2}{c}{\textbf{AndroidControl-Low}} 
& \multicolumn{2}{c}{\textbf{AndroidControl-High}} 
& \multicolumn{2}{c}{\textbf{GUI-Odyssey}} \\
\cmidrule(lr){3-4} \cmidrule(lr){5-6} \cmidrule(lr){7-8}
 &  & \textbf{TM} & \textbf{EM} & \textbf{TM} & \textbf{EM} & \textbf{TM} & \textbf{EM} \\
\midrule
OS-Genesis-7B        & Model          & 90.7 & 74.2 & 65.9 & 44.4 & 11.7 & 3.6 \\
OS-Atlas-7B          & Human\&Model   & 73.0 & 67.3 & 70.4 & 56.5 & 91.8* & 76.8* \\
Aguvis-7B            & Human\&Model   & 93.9 & \underline{89.4} & 65.6 & 54.2 & 26.7 & 13.5 \\
\midrule
Qwen2.5-VL-7B        & --             & 94.1 & 85.0 & 75.1 & 62.9 & 59.5 & 46.3 \\
Ours (aw-app)        & Model          & \textbf{95.4} & \textbf{90.2} & \underline{76.1} & \textbf{66.1} & \textbf{65.8} & \textbf{48.4} \\
Ours (extra-app)     & Model          & \underline{95.3} & 89.3 & \textbf{76.3} & \underline{65.2} & \underline{65.4} & \underline{47.6} \\
\bottomrule
\end{tabular}
}
\caption{Performance comparison on the AndroidControl and GUI-Odyssey benchmarks. Results are reported without the open action type. *OS-Atlas employs different train/test splits on GUI-Odyssey and is thus not directly comparable. \textbf{Bold} and \underline{underline} indicate the best and second-best results. }
\label{tab:sft_results}
\end{table*}

\subsection{Experiment Settings}\label{sec:setup}

\paragraph{Benchmarks.}
We use two widely used benchmarks, the AndroidControl~\cite{Li2024OnTE} and GUI-Odyssey~\cite{Lu2024GUIOA}, to evaluate the quality of our synthesized data. We report Type Match (TM) and Exact Match (EM) as the primary evaluation metrics. To evaluate the effectiveness of our~\name for trajectory completion verification, we further construct an evaluation set from execution traces in AndroidWorld~\cite{Rawles2024AndroidWorldAD}. Specifically, we collect more than $300$ trajectories containing both positive and negative examples, and report Accuracy and F1 score. To evaluate the effectiveness of the overall~\name framework, we construct a benchmark consisting of $86$ queries with their corresponding initial environment snapshots and annotated reference goal states. Half of the queries are used as the training split. We report the Success Rate (SR) on our self-built benchmark, with results verified through both manual assessment and~\name.

\paragraph{Baselines.}
To evaluate the quality of the synthesized data, we selected three advanced open-source models, Os-Genesis~\cite{Sun2024OSGenesisAG}, OS-Atlas~\cite{Wu2024OSATLASAF}, and Aguvis~\cite{Xu2024AguvisUP}, for comparison with our fine-tuned model. Moreover, we compare our method with three representative model-based approaches for reward verification from prior work: DigiRL~\cite{Bai2024DigiRLTI}, DistRL~\cite{Wang2024DistRLAA}, and StepCritic~\cite{Lai2025AndroidGenBA}. Additionally, we include two baselines that use only the goal-state screenshot and only the anchored goal-state screenshot as input, denoted as GS-Only and GSA-Only.

\paragraph{Training Details.}  
For the supervised fine-tuning (SFT) experiments on synthesized trajectory data, we adopt Qwen2.5-VL-7B~\cite{Bai2025Qwen25VLTR} as the base model and fine-tune it using LoRA~\cite{Hu2021LoRALA}. For model-based reward evaluation, we use several vision-language models as judge models, including Qwen3-VL-8B, Qwen3-VL-32B~\cite{Yang2025Qwen3TR}, GPT-4o~\cite{Hurst2024GPT4oSC}, and Gemini-2.5-Pro~\cite{Comanici2025Gemini2P}. In the online reinforcement learning (RL) experiments, we adopt UI-TARS-7B-DPO~\cite{Qin2025UITARSPA} and GUI-Owl-7B~\cite{Ye2025MobileAgentv3FA} as the backbone models, and use GRPO~\cite{Shao2024DeepSeekMathPT} as the optimization algorithm, with Qwen3-VL-32B serving as the judge model. Detailed experimental settings are provided in Appendix~\ref{appendix:training}.

\begin{table*}[ht]
\centering
\resizebox{\linewidth}{!}{
\setlength{\tabcolsep}{10pt}
\begin{tabular}{l c c c c c c c c c c}
\toprule
\multirow{2}{*}{\textbf{Method}} 
& \multicolumn{2}{c}{\textbf{Qwen3-VL-8B}} 
& \multicolumn{2}{c}{\textbf{Qwen3-VL-32B}} 
& \multicolumn{2}{c}{\textbf{GPT-4o}} 
& \multicolumn{2}{c}{\textbf{Gemini-2.5-pro}}
& \multicolumn{2}{c}{\textbf{Avg}}\\
\cmidrule(lr){2-3} \cmidrule(lr){4-5} \cmidrule(lr){6-7} \cmidrule(lr){8-9} \cmidrule(lr){10-11}
 & \textbf{Acc} & \textbf{F1} & \textbf{Acc} & \textbf{F1} & \textbf{Acc} & \textbf{F1} & \textbf{Acc} & \textbf{F1} & \textbf{Acc} & \textbf{F1}\\
\midrule
DigiRL       & 67.0 & 64.9 & 71.4 & 65.9 & 63.8 & 56.2 & 71.1 & 64.3 & 68.3 & 62.8 \\
DistRL       & 67.9 & 69.3 & 80.0 & 79.7 & 69.8 & 66.9 & 82.5 & 82.3 & 75.0 & 74.5 \\
StepCritic   & 64.4 & 74.1 & 76.8 & 81.0 & 81.3 & 84.0 & 87.9 & 88.3 &77.6 & 81.8 \\
\midrule
GS-Only & 84.1 & 84.2 & 83.8 & 83.9 & 78.1 & 78.8 & 86.0 & 85.9 &83.0 & 83.2 \\
GSA-Only  & 87.6 & 88.2 & 86.4 & 86.9 & 81.3 & 80.9 & 89.8 & 90.1 &86.3 & 86.5 \\
\name     & \textbf{90.2} & \textbf{91.0} & \textbf{92.1} & \textbf{92.4} & \textbf{91.1} & \textbf{91.3} & \textbf{92.7} & \textbf{92.7} & \textbf{91.5} & \textbf{91.8} \\
\bottomrule
\end{tabular}
}
\caption{Offline trajectory evaluation results of different model-based reward methods.}
\label{tab:offline_reward_results}
\end{table*}

\subsection{Main Results}
We first evaluate the fine-tuned model to validate data quality, then assess the accuracy of ~\name on offline trajectories, and finally verify that the entire pipeline scales effectively to online RL.

\paragraph{Evaluation of Synthesized Data Quality.}
We evaluate the effectiveness and scalability of synthesized data for GUI agent training. The model is fine-tuned on data collected from 20 AndroidWorld apps and further scaled using data from 40 selected open-source apps.
As shown in Table~\ref{tab:sft_results}, fine-tuning on synthesized data leads to consistent improvements across both benchmarks, with notable gains on the EM metric for the low and high splits of AndroidControl (+5.2\% and +3.2\%, respectively) and on the TM metric for GUI-Odyssey (+6.3\%). When scaling to additional open-source apps, the model achieves similarly strong performance, although slightly lower due to distribution differences between datasets. These results indicate that \textbf{our synthesized data effectively improves model performance and generalizes well across diverse application domains.} Additional results on other base models are provided in Appendix~\ref{appendix:supplementary_sft}.

\paragraph{Offline Evaluation for \name.}
We conduct offline evaluations to assess the ability of~\name in judging task completion, comparing it with three model-based reward methods as shown in Table~\ref{tab:offline_reward_results}. As model capabilities continue to improve, the accuracy and F1 scores of the three baseline methods also increase. However, they still lag behind the rule-based approach, whose accuracy and F1 are theoretically 100\%. 
Providing the model with a reference state (GS-Only or GSA-Only) significantly enhances its evaluation capability. \name achieves the highest results in evaluating trajectory correctness, being the only method to surpass 90\% in both accuracy and F1 for both open-source and closed-source models. This demonstrates that \textbf{\name most closely approximates rule-based evaluation while retaining scalability advantages.}

\paragraph{RL Training with the Full Framework.}
\begin{figure}[t]
    \centering
    \begin{subfigure}[b]{\linewidth}
        \centering
        \includegraphics[width=\linewidth]{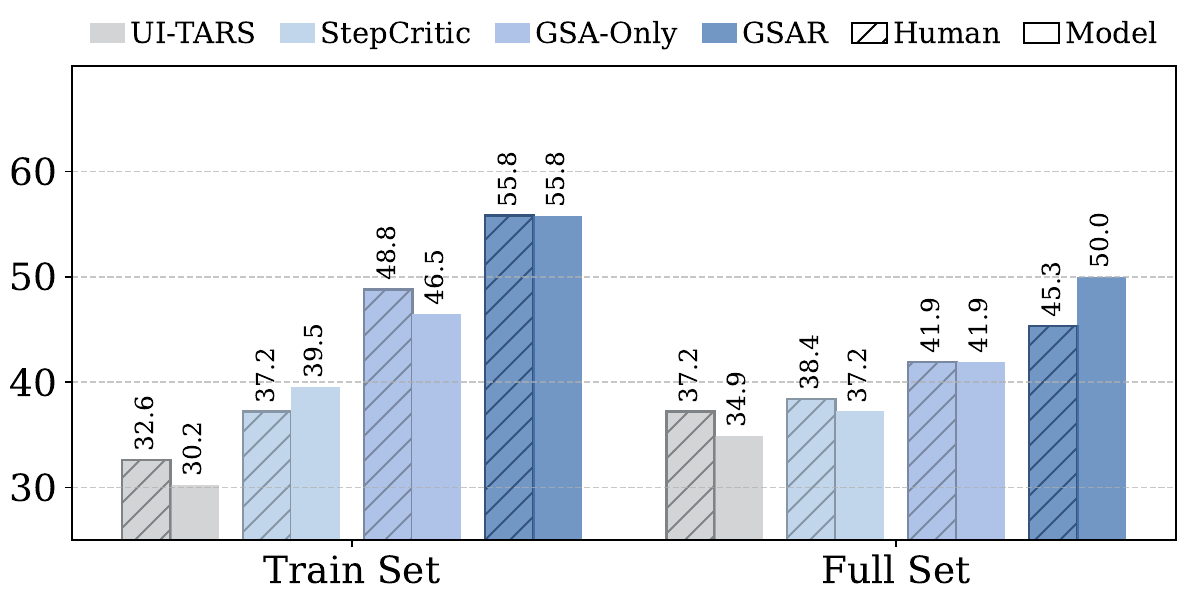}
        \caption{UI-TARS-7B-DPO}
        \label{fig:online_results_subfig_a}
    \end{subfigure}
    \begin{subfigure}[b]{\linewidth}
        \centering
        \includegraphics[width=\linewidth]{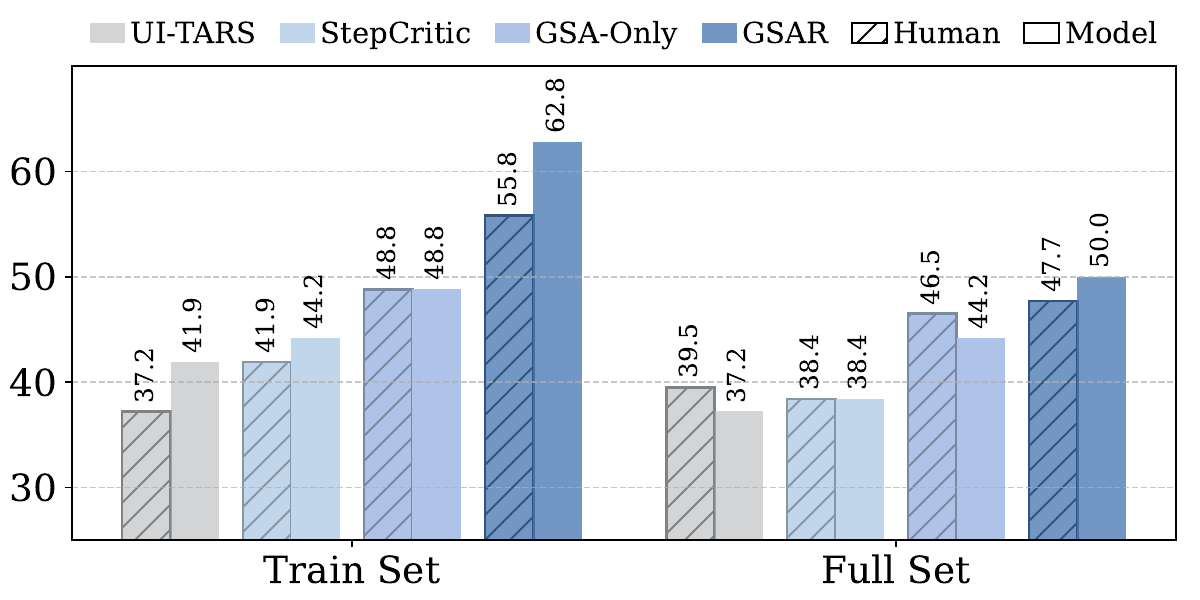}
        \caption{GUI-Owl-7B}
        \label{fig:online_results_subfig_b}
    \end{subfigure}

\caption{Success rate on the self-built benchmark. \textit{Model} indicates that GSAR is used as the evaluation method.}
\label{fig:online_results}
\end{figure}

To further assess the effectiveness and efficiency of~\name, we use our self-built benchmark, including triplets (task, initial environment, goal-state-anchor reference), for online RL. As shown in Figure~\ref{fig:online_results}, the model trained with \name as the reward function achieves a 23.2\% improvement on the training split and an 8.1\% improvement on the full task set compared to the UI-TARS. A similar trend is observed for GUI-Owl, with gains of 18.6\% and 8.2\%, respectively. Moreover, incorporating the goal-state-anchor reference substantially enhances task execution over StepCritic, while adding action history also improves training compared to GSA-Only. Verification using GSAR shows similar trends to human evaluation, with some metrics closely matching human judgments. Overall, \textbf{the proposed pipeline effectively enables scalable RLVR training in GUI agents and improves their performance.}

\begin{figure}[t]
  \centering

  \begin{subfigure}[t]{0.49\linewidth}
    \centering
    \includegraphics[width=\linewidth]{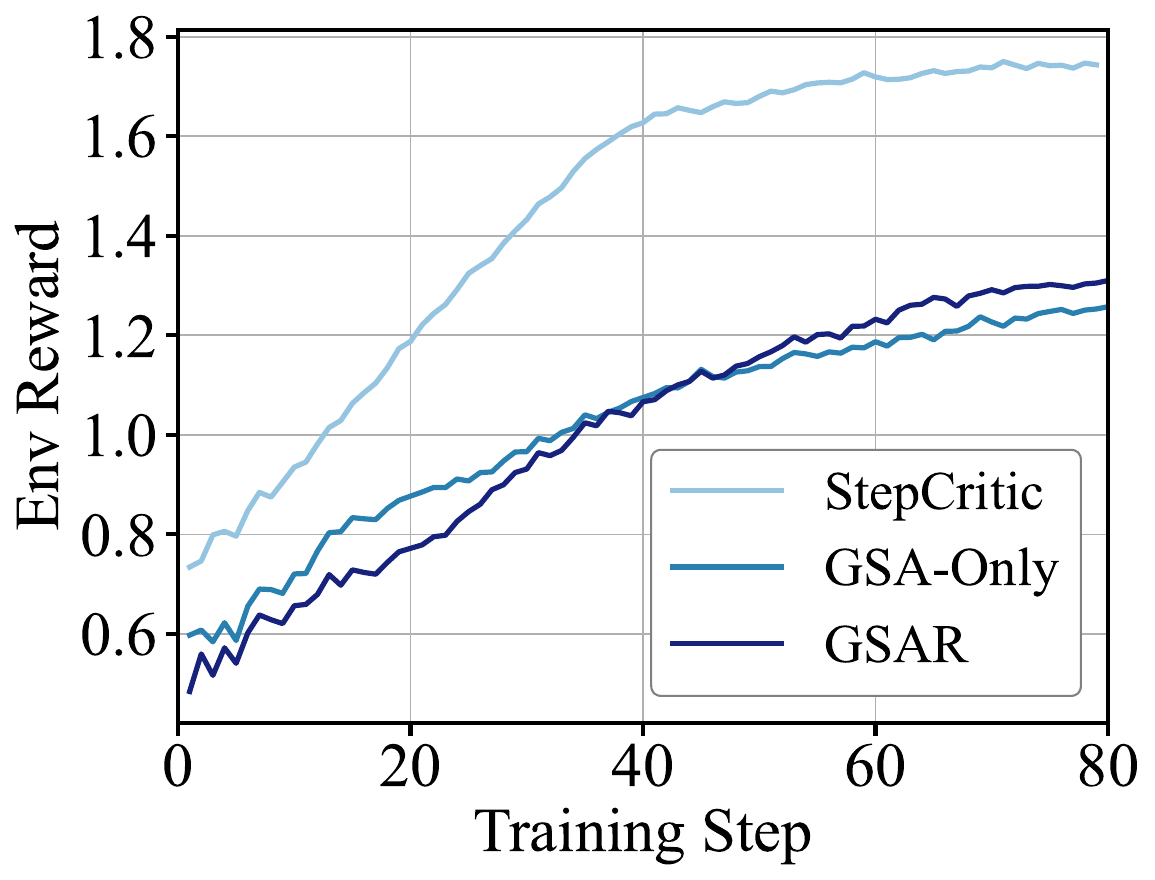}
    \caption{Reward on UI-TARS.}
    \label{fig:reward_curve_uitars}
  \end{subfigure}
  \hfill
  \begin{subfigure}[t]{0.49\linewidth}
    \centering
    \includegraphics[width=\linewidth]{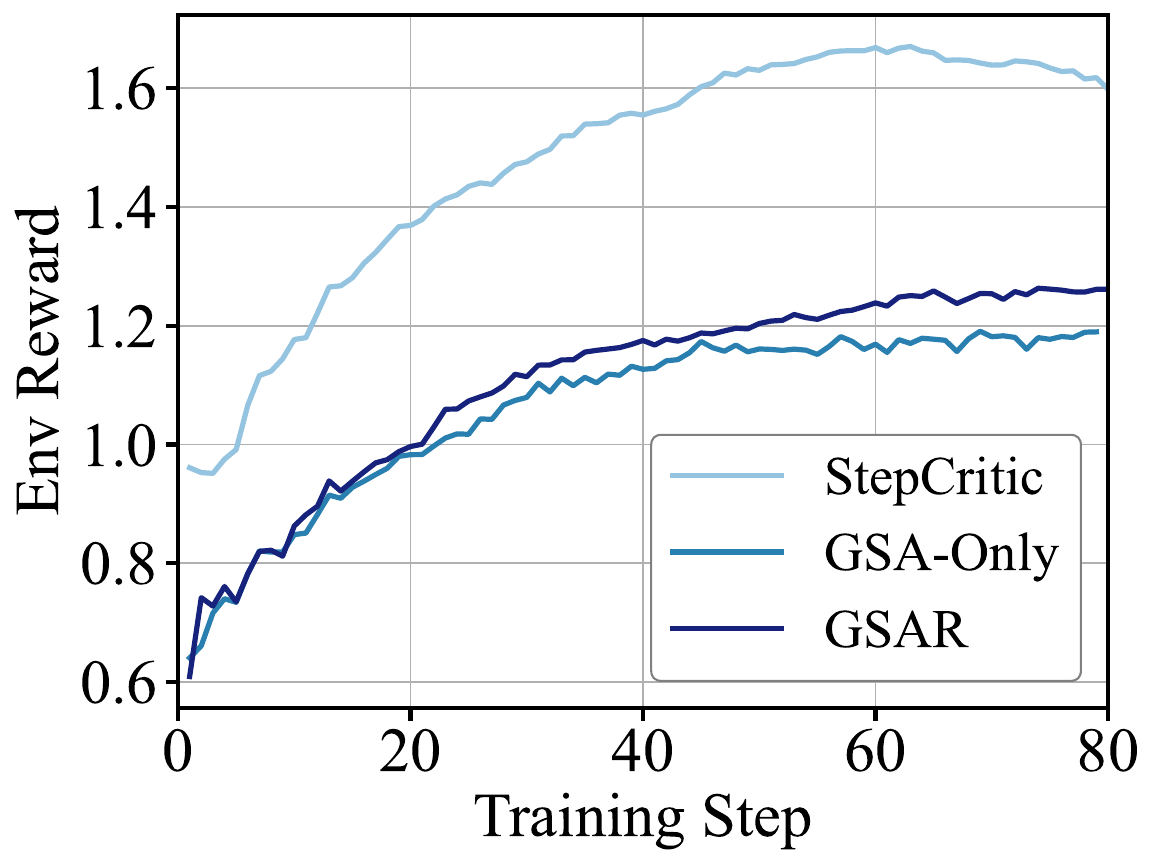}
    \caption{Reward on GUI-Owl.}
    \label{fig:reward_curve_guiowl}
  \end{subfigure}

  \caption{Online training reward curves of three reward mechanisms on the UI-TARS-7B-DPO and GUI-Owl-7B models. Each curve represents the per-step reward across training steps.}
  \label{fig:reward_curves}
\end{figure}

\subsection{Ablation Study.}

We perform ablation experiments on offline verification to assess the contribution of each component. Specifically, we test three variants: (1) w/o history, which removes the action history; (2) w/o anchor, which removes the anchor on the goal-state screenshot; and (3) w/o reference, which removes the goal-state screenshot.
\begin{table}[t]
\centering
\resizebox{\linewidth}{!}{
\setlength{\tabcolsep}{10pt}
\begin{tabular}{l c c c c}
\toprule
\multirow{2}{*}{\textbf{Method}} 
& \multicolumn{2}{c}{\textbf{Qwen3-VL-8B}} 
& \multicolumn{2}{c}{\textbf{Qwen3-VL-32B}} \\
\cmidrule(lr){2-3} \cmidrule(lr){4-5}
& \textbf{Acc} & \textbf{F1} & \textbf{Acc} & \textbf{F1} \\
\midrule
w/o reference & 64.4 & 74.1  & 76.8 & 81.0  \\
w/o anchor & 87.3 & 88.0 & 90.5 & 90.7  \\
w/o history & 87.6 & 88.2  & 86.4 & 86.9 \\
\name(Ours) & \textbf{90.2} & \textbf{91.0} & \textbf{92.1} & \textbf{92.4} \\
\bottomrule
\end{tabular}
}
\caption{The results of the ablation study.}
\label{tab:ablation_results}
\end{table}

As shown in Table~\ref{tab:ablation_results}, \textbf{combining the action history with the anchored reference screenshot as input yields the best performance, and removing either component leads to a noticeable degradation.} Omitting the reference image leads to substantially lower accuracy compared to the rule-based approach, as the model loses information about the final task state. Excluding the anchor annotations from the completed reference also degrades performance, since VLMs may fail to perceive subtle visual differences between similar incomplete and completed screenshots. Additionally, because some tasks involve critical intermediate steps, discarding the action history removes this contextual information, negatively affecting the final decision.

\section{Analysis Experiments}

\subsection{Annotation Accuracy by Task Category}
\begin{table}[t!]
\centering
\resizebox{\linewidth}{!}{
\begin{tabular}{l c c c c}
\toprule
\textbf{Type} & \textbf{Answer} & \textbf{Delete} & \textbf{Normal} & \textbf{Total} \\
\midrule
Count        & 20 & 15 & 47 & 82 \\
Accuracy     & 90.0 & 100.0 & 89.4 & 91.5 \\
\bottomrule
\end{tabular}
}
\caption{Automatic annotation accuracy of goal-state anchoring across task categories.}
\label{tab:annotation_accuracy}
\end{table}

Table~\ref{tab:annotation_accuracy} summarizes the accuracy of automatic goal-state annotation via GPT-4o across task categories. The overall annotation accuracy reaches 91.5\%. Delete tasks achieve 100\% accuracy because the deleted object no longer appears in the goal state, eliminating the need for fine-grained localization; therefore, these cases are treated as fully correct. For Answer and Normal tasks, accuracy is slightly lower, as the goal state may contain multiple small or scattered key elements that the model must individually identify and localize, posing a greater challenge. Overall, the automatic annotation pipeline provides reliable annotations across different task categories and can substantially reduce the need for manual annotation.

\begin{figure}[t]
  \centering
  \includegraphics[width=\linewidth]{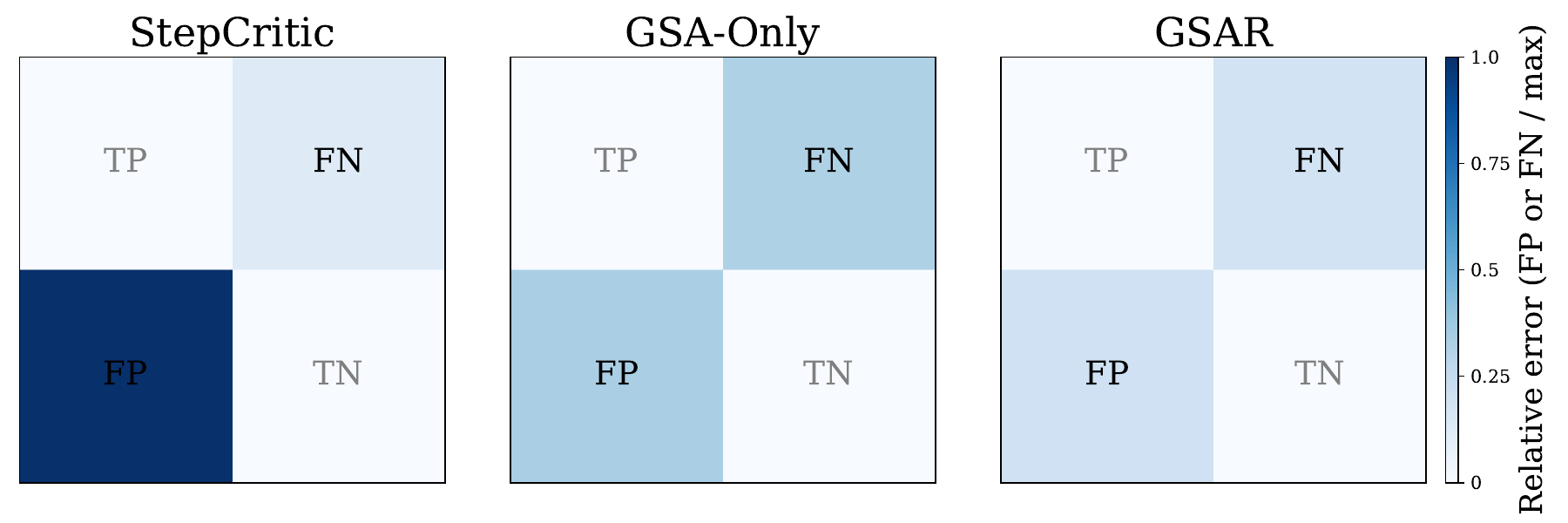}
  \caption{Comparison of false positives (FP) and false negatives (FN) across different reward designs. StepCritic reduces FN but leaves FP high, GSA-Only reduces FP but some FN remain, and combining both mechanisms balances the reward signal while mitigating both FP and FN errors.}
  \label{fig:false_positive}
\end{figure}

\subsection{Understanding the Role of Reward Signals in RL Training}

We investigate the impact of reward signals on reinforcement learning by comparing the performance of models trained with different reward functions. Although StepCritic exhibits relatively low false negative rates in offline evaluation, it suffers from a notably high false positive rate (Figure~\ref{fig:false_positive}). As a result, its reward signals during RL training are overly optimistic (Figure~\ref{fig:reward_curves}), leading to policies that fail to consistently translate into strong online performance and producing agents that are less stable and less reliable than ours (Figure~\ref{fig:online_results}). This suggests that providing more accurate reward signals can lead to greater gains in RL training.

\subsection{Comparison with Rule-based Methods}

\begin{figure}[t]
  \centering
  \begin{subfigure}[t]{0.48\linewidth}
    \centering
    \includegraphics[width=\linewidth]{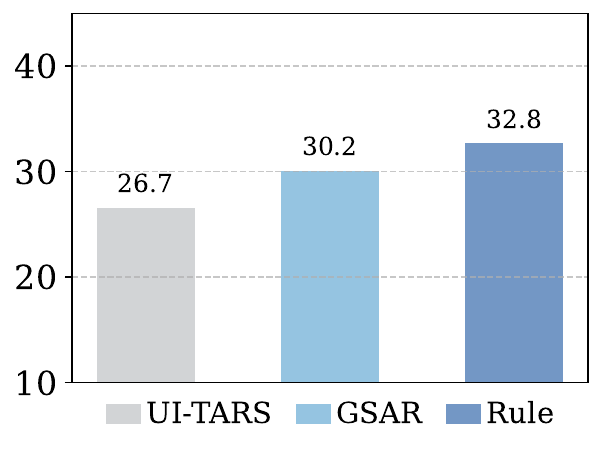}
    \caption{Evaluation results}
    \label{fig:online_bar_chart_uitars_aw}
  \end{subfigure}
  % \hfill
  \begin{subfigure}[t]{0.47\linewidth}
    \centering
    \includegraphics[width=\linewidth]{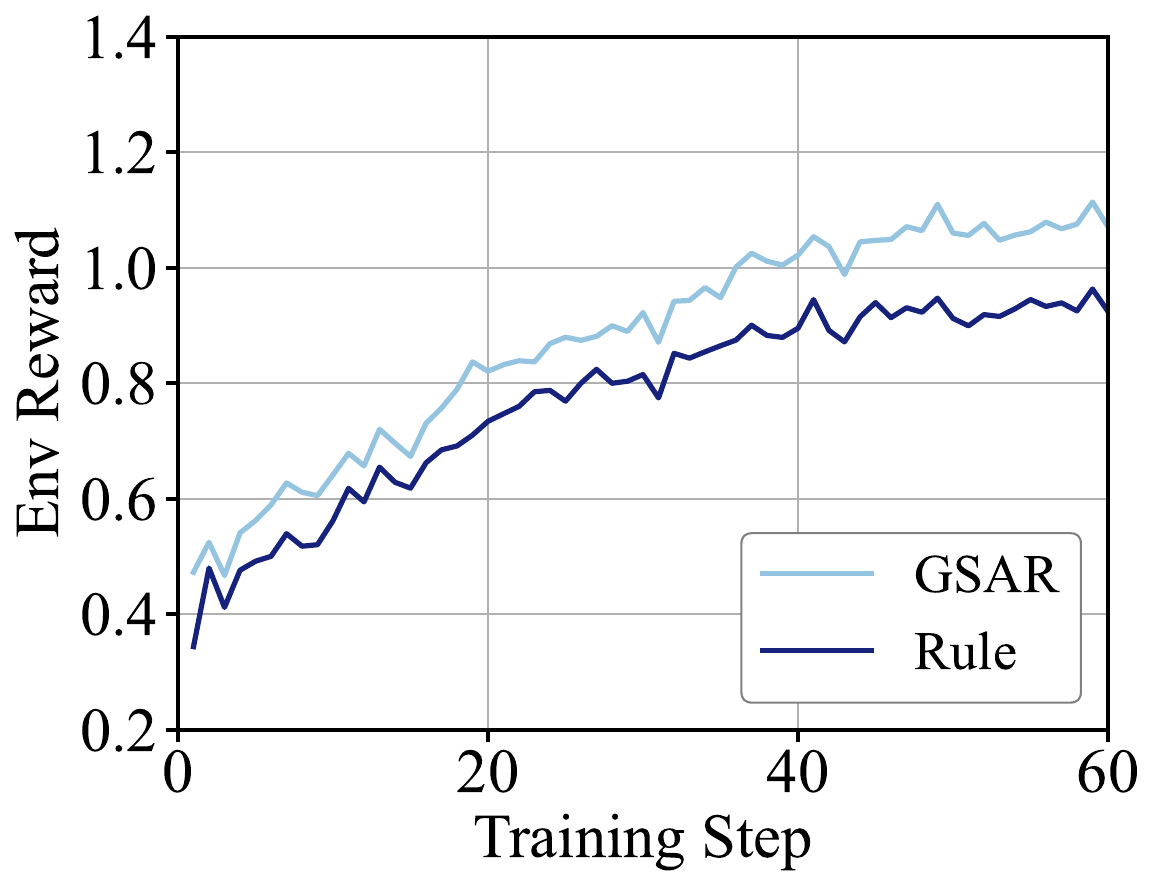}
    \caption{Reward curves}
    \label{fig:reward_curve_uitars_aw}
  \end{subfigure}

  \caption{Evaluation results on AndroidWorld and online training reward curves of the UI-TARS model.}
  \label{fig:uitars_aw}
\end{figure}

\paragraph{Performance Comparison.}
We conduct a comparative study between GSAR and rule-based rewards on AndroidWorld~\cite{Rawles2024AndroidWorldAD}. We select a subset of 42 tasks as the training set, while evaluation is conducted on the full benchmark. As shown in Figure~\ref{fig:online_bar_chart_uitars_aw}, both GSAR and rule-based rewards improve the success rate over the UI-TARS-7B-DPO baseline (26.7\%). GSAR increases the performance to 30.2\% (+3.5\%), while rule-based rewards achieve 32.8\% (+6.1\%) due to their near-noiseless verification signals. Meanwhile, the reward curves in Fig.~\ref{fig:reward_curve_uitars_aw} show that GSAR follows a trend highly consistent with rule-based rewards, demonstrating that it can provide stable optimization signals without relying on manually designed rules, offering better scalability and generalization.

\begin{table}[t!]
\centering
\setlength{\tabcolsep}{11pt}
\resizebox{\linewidth}{!}{
\begin{tabular}{l c}
\toprule
\textbf{Reward Method} & \textbf{Average Rollout Time (s)} \\
\midrule
Rule-based & 2754.92 \\
GSAR (Model-based) & \textbf{2342.02} \\
\bottomrule
\end{tabular}
}
\caption{Average rollout time per training step for different reward methods.}
\label{tab:rollout_time}
\end{table}

\paragraph{Rollout Efficiency.}
Both methods perform reward evaluation on a step-wise basis during RL training. GSAR introduces additional latency due to model-based evaluation after each action. However, for most tasks, the latency introduced by model inference is lower than the overhead incurred by repeated ADB-based device access for rule-based verification. As shown in Table~\ref{tab:rollout_time}, GSAR therefore achieves lower overall rollout latency than the rule-based baseline.
\section{Conclusion}
We propose \name (Goal-State-Anchor Reward), a framework that integrates self-evolving data synthesis with a state-anchor mechanism. To enhance environmental diversity during data synthesis, we introduce an iterative mechanism that evolves environments through task execution and incorporate task complexification to generate diverse tasks. Furthermore, we leverage final states of successful trajectories to anchor task-relevant UI elements, providing richer context for model-based reward evaluation. Our experiments demonstrate that \name achieves over 90\% accuracy in trajectory verification and improves performance over the base model in online RL training. Moreover, the reward curve of \name closely aligns with the rule-based reward, demonstrating its ability to provide stable optimization signals without relying on handcrafted rules. Overall, \name provides a scalable data synthesis paradigm and provides a reliable reward framework for training GUI agents.

\section*{Limitations}
Our method addresses the scalability issues of rule-based approaches and the accuracy limitations of model-based approaches. However, it considers only a single final state as the success criterion for each task, whereas a task may be achievable through multiple alternative solutions. Moreover, for question-answering tasks, our method may assign a correct reward before the actual correct answer is produced. For tasks whose completion cannot be fully expressed visually (e.g., deleting certain items), it is difficult for the model to annotate reasonable task-relevant elements, and natural language or other modalities may be necessary to describe the final state. Future work should involve a more fine-grained categorization and handling of task types during data collection and training.

Meanwhile, both the automated task execution for trajectory collection and the filtering process are performed by VLMs (e.g., GPT-4o), eliminating the need for manual annotation. However, the quality of the generated trajectories is sometimes constrained by the capabilities of the models. For tasks that cannot be completed automatically, manual execution is still required to obtain the correct goal state if they are to be used for reinforcement learning. We expect that stronger GUI-capable models in the future will further improve the utilization of model-generated data.

\bibliography{acl2026}

\appendix
\newpage

\section{Benchmarks}
\label{appendix:benchmarks}
Here we provide additional details on the benchmarks used to evaluate~\name.

\paragraph{AndroidControl.}
AndroidControl~\cite{Li2024OnTE} is a large-scale dataset for human-computer interface control on Android devices, containing 15,283 task demonstrations across 833 apps. Each task includes both high-level and low-level human-generated instructions, covering 14,548 unique tasks. The dataset enables analysis of agent performance on tasks of varying complexity, both within the training domain and out-of-domain. AndroidControl supports studies on how fine-tuning with more data affects performance, particularly highlighting the challenges of generalizing to unseen apps or higher-level tasks.

\paragraph{GUI-Odyssey.}
GUI-Odyssey~\cite{Lu2024GUIOA} is a large-scale dataset for cross-app mobile GUI navigation, consisting of 8,334 episodes with an average of 15.3 steps per episode. It spans 6 mobile devices, 212 apps, and 1,357 app combinations. Each step includes detailed semantic reasoning annotations to support models in complex cross-app reasoning and decision-making. The dataset enables training and evaluation of agents capable of long-step, multi-app navigation tasks.

Following~\cite{Zhang2025AgentCPMGUIBM}, we removed steps related to the “open\_app” action type in the evaluation of these benchmarks.

\section{Self-Evolving Data Synthesis}
\label{appendix:self_evolving}
In this section, we present the algorithmic workflow of the self-evolving data synthesis framework used for generating tasks and collecting trajectories in mobile GUI environments. The framework iteratively collects tasks, evolves the GUI environment, and generates increasingly complex instructions based on previously collected trajectories. The workflow, summarized in Algorithm~\ref{alg:self_evolving_gui}, illustrates how the system produces a diverse and high-quality dataset suitable for reinforcement learning training and goal-state–anchored reward evaluation.

\begin{algorithm}[ht]
\caption{Self-Evolving Data Synthesis}
\begin{algorithmic}[1]
\Require 
    Device serial \(d\), package name \(p\), maximum iterations \(T\),\\
    maximum tasks per page \(M_t\), maximum breadth \(M_b\), maximum depth \(M_d\),\\
    maximum steps per task \(S_{\max}\)

\Ensure 
    Task set \( \mathcal{Q} \) and trajectory set \( \mathcal{D} \)

\State \textbf{Initialize} task set \( \mathcal{Q} \gets \emptyset \)
\State \textbf{Initialize} trajectory set \( \mathcal{D} \gets \emptyset \)
\State \textbf{Initialize} previous trajectory set \( \mathcal{D}_{prev} \gets \emptyset \)

\For{iteration \( t = 1 \dots T \)}

    \State \textbf{Pull} environment \(E_t\) from device

    \State \textbf{Collect} candidate tasks 
    \( \mathcal{Q}_t = \textsc{CollectTask}(E_t, M_t, M_b, M_d) \)

    \If{\(t > 1\) and \( \mathcal{D}_{prev} \neq \emptyset \)}
        \For{each trajectory \( \tau \in \mathcal{D}_{prev} \)}
            \If{\( \tau \) is completed and not complexified}
                \State \textbf{Generate} complexified tasks
                \( \mathcal{Q}' = \textsc{ComplexifyTask}(\tau, E_t) \)
                \State \( \mathcal{Q}_t \gets \mathcal{Q}_t \cup \mathcal{Q}' \)
            \EndIf
        \EndFor
    \EndIf

    \State \textbf{Update} task set \( \mathcal{Q} \gets \mathcal{Q} \cup \mathcal{Q}_t \)

    \State \textbf{Push} environment \(E_t\) back to device

    \State \textbf{Execute} tasks \( \mathcal{Q}_t \) with step limit \(S_{\max}\)

    \State \textbf{Collect} resulting trajectories \( \mathcal{D}_t \)

    \State \textbf{Update} trajectory set \( \mathcal{D} \gets \mathcal{D} \cup \mathcal{D}_t \)

    \State \textbf{Set} \( \mathcal{D}_{prev} \gets \mathcal{D}_t \)

\EndFor

\State \Return \( \mathcal{Q}, \mathcal{D} \)

\end{algorithmic}
\label{alg:self_evolving_gui}
\end{algorithm}

We use GPT-4o~\cite{Hurst2024GPT4oSC} for task generation, task complexification, and task execution. For AndroidWorld apps, we set the maximum iteration to 6. For additional open-source apps, we first allow the model to perform a single random exploration to generate trajectories, and then count the number of unique pages in the trajectory screenshots, mapping this number to a range of 1–6 to determine the maximum number of self-evolving iterations for that app. This approach ensures that apps with more diverse interactions undergo more iterations, while avoiding unnecessary exploration time for simpler apps.

\section{Details of Experiments}
\label{appendix:training}

\paragraph{Supervised Fine-Tuning.}
We provide the detailed experimental settings for the supervised fine-tuning (SFT) stage in Table~\ref{tab:sft_details}, including the training data, model configuration, and key hyperparameters used during training.
For all fine-tuned models, the training dataset contains 6,331 steps, covering both low-level and high-level instruction types, where the low-level instructions correspond to the action reasoning output.

\begin{table}[h]
\centering
\footnotesize
\setlength{\tabcolsep}{14pt}
\resizebox{\linewidth}{!}{
\begin{tabular}{lc}
\toprule
\textbf{Hyperparameters} & \textbf{All methods} \\
\midrule
Fine-tuning method & LoRA \\
LoRA rank (\(r\)) & 8 \\
LoRA \(\alpha\) & 16 \\
Train batch size & 128 \\
Training epochs & 3 \\
Learning rate & \(1\times10^{-5}\) \\
LR Scheduler & Cosine \\
Warmup Ratio & 0.1 \\
GPU numbers & 8 \\
\bottomrule
\end{tabular}
}
\caption{Training hyperparameters for SFT.}
\label{tab:sft_details}
\end{table}

\paragraph{Offline Verification for \name.}
The datasets used to evaluate~\name and other methods are derived from the evaluation trajectories of GUI-Owl-7B and Qwen2.5-VL-7B on the AndroidWorld benchmark. To better assess the differences between methods, we constructed a total of 315 trajectories, including 164 positive samples and 151 negative samples. For each task, the goal state is selected from the last step screenshot and anchored using GPT-4o. To ensure fairness, we strictly follow the experimental settings and prompts described in the original papers when using these baselines.

\begin{table}[h]
\centering
\footnotesize
\setlength{\tabcolsep}{14pt}
\resizebox{\linewidth}{!}{
\begin{tabular}{lc}
\toprule
\textbf{Hyperparameters} & \textbf{All methods} \\
\midrule
Train batch size & 32 \\
PPO batch size & 256 \\
Training steps & 80 \\
Max turn & 20 \\
Rollout numbers & 8 \\
Temperature & 0.7 \\
Learning rate & \(1\times10^{-6}\) \\
KL coefficient & 0.001 \\
Clip ratio & 0.2 \\
\bottomrule
\end{tabular}
}
\caption{Training hyperparameters for RL.}
\label{tab:rl_details}
\end{table}

\begin{table*}[t!]
\centering
\resizebox{\linewidth}{!}{
\begin{tabular}{l@{\hspace{10pt}}cccccc@{\hspace{20pt}}c}
\toprule
\multirow{2}{*}{\textbf{Model}} & \multirow{2}{*}{\textbf{Data Sources}}
& \multicolumn{2}{c}{\textbf{AndroidControl-Low}}
& \multicolumn{2}{c}{\textbf{AndroidControl-High}}
& \multicolumn{2}{c}{\textbf{GUI-Odyssey}} \\
\cmidrule(lr){3-4} \cmidrule(lr){5-6} \cmidrule(lr){7-8}
 &  & \textbf{TM} & \textbf{EM} & \textbf{TM} & \textbf{EM} & \textbf{TM} & \textbf{EM} \\
\midrule
% SFT
UI-TARS-7B-SFT        & --          & \textbf{95.05} & \underline{91.52} & 79.10 & 71.21 & 80.71 & 66.98 \\
\name (aw-app)            & Model       & 94.99 & \textbf{91.53} & \underline{80.33} & \underline{71.97} & \underline{81.78} & \textbf{68.01} \\
\name (extra-app)         & Model       & \underline{95.01} & 91.48 & \textbf{80.76} & \textbf{72.41} & \textbf{81.90} & \underline{67.93} \\
\midrule
% DPO
UI-TARS-7B-DPO        & --          & 92.05 & 88.74 & 80.77 & 73.56 & 85.29 & 68.33 \\
\name (aw-app)            & Model       & \underline{92.82} & \underline{89.52} & \underline{81.04} & \underline{73.85} & \underline{85.56} & \textbf{71.28} \\
\name (extra-app)         & Model       & \textbf{93.33} & \textbf{90.01} & \textbf{81.28} & \textbf{74.09} & \textbf{85.90} & \underline{71.15} \\
\bottomrule
\end{tabular}
}
\caption{Performance comparison of UI-TARS-7B-SFT and UI-TARS-7B-DPO fine-tuned on our synthesized data, evaluated on the AndroidControl and GUI-Odyssey benchmarks. Results are reported without the open action type. \textbf{Bold} and \underline{underline} indicate the best and second-best results within each block.}
\label{tab:sft_results_uitars}
\end{table*}

\paragraph{Online Reinforcement Learning.}
The experimental configurations for the reinforcement learning (RL) stage are summarized in Table~\ref{tab:rl_details}, including the backbone agent, reward evaluation model, optimization algorithm, and other key training parameters used during online RL. The experiments were conducted on two nodes, each equipped with \(8 \times \text{NVIDIA H100 GPUs}\). 

To support large-scale environment interactions while decoupling the GUI environment from the training process, we deployed a service of 128 Android emulators for parallel trajectory sampling. Specifically, each of the two machines launched 64 emulators, providing a total of 128 emulators that enable the agent to interact with the environment independently of the training, thereby facilitating efficient training and data collection.

\section{Supplementary SFT Results}
\label{appendix:supplementary_sft}

To further verify the generalization of our synthesized data across different base models, we fine-tune both UI-TARS-7B-SFT and UI-TARS-7B-DPO~\cite{Qin2025UITARSPA} on the same training data, as shown in Table~\ref{tab:sft_results_uitars}. Across all three benchmarks, our method performs on par with, and in several cases surpasses, the strong UI-TARS baseline, and scaling to additional open-source apps yields comparable results. We note that the margins of improvement on UI-TARS are narrower than those observed on Qwen2.5-VL-7B~\cite{Bai2025Qwen25VLTR} (Table~\ref{tab:sft_results}), likely because UI-TARS is already specialized for GUI tasks and achieves substantially stronger baseline performance, leaving less room for further gains. These results confirm that the quality of our synthesized data is not tied to a specific base model, and the benefits transfer across different model initializations and capabilities.

\section{Additional Analysis Results}
\label{appendix:additional_analysis}

\subsection{Analyzing the Effectiveness of Self-Evolving Data Synthesis}
\begin{table}[t]
\centering
\resizebox{\linewidth}{!}{
\begin{tabular}{lccc}
\toprule
\textbf{Stage} & \textbf{Edge Density} & \textbf{Entropy} & \textbf{UI Elements} \\
\midrule
Early (Iter 1--2) & 0.0240 & 1.90 & 22.68 \\
Later (Iter 5--6) & \textbf{0.0295} & \textbf{2.11} & \textbf{24.56} \\
\bottomrule
\end{tabular}
}
\caption{Impact of self-evolving process on GUI page complexity.}
\label{tab:self_evolving}
\end{table}

\begin{table}[t]
\centering
\resizebox{\linewidth}{!}{
\begin{tabular}{lcc}
\toprule
\textbf{Task Type} & \textbf{Task string length} & \textbf{Trajectory length} \\
\midrule
Normal & 144.41 & 7.91 \\
Complexified & \textbf{338.92} & \textbf{12.08} \\
\bottomrule
\end{tabular}
}
\caption{Effect of task complexity on average query string length and trajectory length.}
\label{tab:task_complexity}
\end{table}

\begin{table}[t]
\centering
\resizebox{\linewidth}{!}{
\begin{tabular}{lcccccc}
\toprule
\textbf{Dataset} & \textbf{$R$=1} & \textbf{$R$=2} & \textbf{$R$=3} & \textbf{$R$=4} & \textbf{$R$=5} & \textbf{$R \geq 4$ Ratio} \\
\midrule
aw-app & 44 & 281 & 214 & 126 & 328 & 45.7\% \\
extra-app & 110 & 342 & 196 & 83 & 229 & 32.5\% \\
\bottomrule
\end{tabular}
}
\caption{Data utilization statistics of synthesized trajectories.}
\label{tab:data_utilization}
\end{table}

\paragraph{Impact of Self-Evolving Mechanism.}
To evaluate the effect of the self-evolving mechanism on page complexity, we sample 300 tasks each from early iterations and later iterations, as summarized in Table~\ref{tab:self_evolving}. For each task, one state is randomly sampled from its trajectory to compute page complexity metrics. The results show that pages from later iterations exhibit slightly higher edge density, increased image entropy, and more UI elements on average. These observations suggest that the self-evolving process gradually increases the visual and functional complexity of generated pages, leading to richer and more diverse environments.

\paragraph{Effect of Task Complexity.}  
We evaluate the impact of task complexification on a filtered set of 339 samples for each setting, as shown in Table~\ref{tab:task_complexity}. After task complexification, both the average length of natural language task descriptions and the average trajectory length increase, demonstrating the effectiveness of the task complexification process.

\paragraph{Data Utilization Analysis}
We analyze the data utilization rate of trajectories generated by our self-evolving data synthesis pipeline, as shown in Table~\ref{tab:data_utilization}. The aw-app and extra-app datasets contain 993 and 960 tasks, respectively, and trajectories with reward score $R \geq 4$ are used for fine-tuning. The results show that a substantial portion of generated trajectories can be effectively utilized for training. The aw-app dataset achieves a higher utilization rate of 45.7\%, benefiting from the carefully selected AndroidWorld applications with stable and complete functionalities.
In contrast, the extra-app dataset has a lower utilization rate of 32.5\%, since it consists of open-source applications, some of which have incomplete functionality or belong to relatively niche usage scenarios.

\subsection{Offline Evaluation by Task Category}
\begin{table*}[t]
\centering
\resizebox{\linewidth}{!}{
\setlength{\tabcolsep}{16pt}
\begin{tabular}{l c c c c c c c c}
\toprule
\multirow{2}{*}{\textbf{Method}} 
& \multicolumn{2}{c}{\textbf{Delete}} 
& \multicolumn{2}{c}{\textbf{Answer}} 
& \multicolumn{2}{c}{\textbf{Normal}} 
& \multicolumn{2}{c}{\textbf{Overall}} \\
\cmidrule(lr){2-3} \cmidrule(lr){4-5} \cmidrule(lr){6-7} \cmidrule(lr){8-9}
 & \textbf{Acc} & \textbf{F1} & \textbf{Acc} & \textbf{F1} & \textbf{Acc} & \textbf{F1} & \textbf{Acc} & \textbf{F1} \\
\midrule
DigiRL      & 81.4 & 80.0 & 41.4 & 0.0   & 79.0 & 75.5 & 71.1 & 64.3 \\
DistRL      & 86.4 & 86.2 & 75.7 & 80.0  & 83.9 & 82.1 & 82.5 & 82.3 \\
StepCritic  & 96.6 & 96.7 & 77.1 & 77.8  & 89.3 & 89.7 & 87.9 & 88.3 \\
GS-Only     & 91.5 & 92.1 & 77.1 & 79.0  & 87.6 & 86.7 & 86.0 & 85.9 \\
GSA-Only    & 88.1 & 88.5 & 85.7 & 88.1  & \textbf{91.9} & \textbf{91.5}  & 89.8 & 90.1 \\
GSAR        & \textbf{98.3} & \textbf{98.3} & \textbf{90.0} & \textbf{90.9} & \textbf{91.9} & \textbf{91.5} & \textbf{92.7} & \textbf{92.7} \\
\bottomrule
\end{tabular}
}
\caption{Offline trajectory evaluation results of different model-based reward methods across task categories.}
\label{tab:offline_reward_type}
\end{table*}

Table~\ref{tab:offline_reward_type} reports classification results by task category using Gemini-2.5-Pro~\cite{Comanici2025Gemini2P}. GSAR demonstrates consistently strong performance across all task types, highlighting both the varying difficulty of trajectory completion judgment across categories and the effectiveness of GSAR in handling them.

StepCritic outperforms both GSA-Only and GS-Only, indicating that action history plays a critical role in determining completion for Delete type tasks. Notably, GSA-Only performs worse than GS-Only on Delete tasks. This may be attributed to the fact that the added anchor boxes tend to cover large screen regions, causing the model to attend to irrelevant elements and thereby degrading judgment accuracy.

Answer tasks require a combination of semantic understanding and goal state reference for accurate judgment. DigiRL yields an F1 of zero on this task type because its frame similarity filter compares the last two screenshots and labels trajectories with minimal visual change as incomplete. Since Answer tasks typically exhibit little visual difference between the final two frames, DigiRL fails to identify any positive samples, resulting in zero true positives for both precision and recall, and consequently an F1 of zero. Comparing GS-Only and GSA-Only, anchoring task-relevant UI elements on the goal state brings substantial gains for Answer tasks.

For Normal tasks, successful and incomplete trajectories exhibit distinct differences in key UI elements, allowing strong performance to be achieved even without action history.

\begin{figure}[t!]
    \centering
    \begin{subfigure}[b]{0.9\linewidth}
        \centering
        \includegraphics[width=\linewidth]{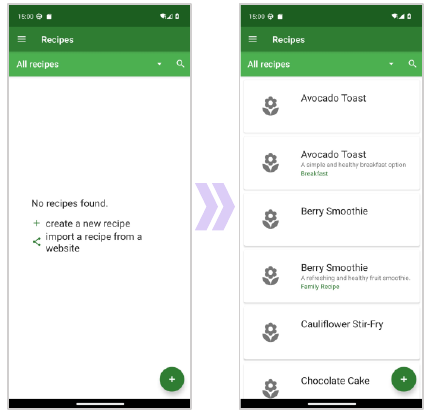}
        \caption{Broccoli}
        \label{fig:case_self_evolve_subfig_a}
    \end{subfigure}
    \begin{subfigure}[b]{0.9\linewidth}
        \centering
        \includegraphics[width=\linewidth]{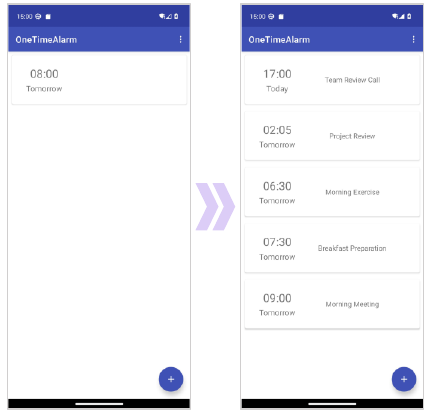}
        \caption{OneTimeAlarm}
        \label{fig:case_self_evolve_subfig_b}
    \end{subfigure}
    \caption{Examples of environment state changes under self-evolving data synthesis.}
    \label{fig:case_self_evolve}
\end{figure}

\begin{figure}[t!]
\centering
\includegraphics[width=\linewidth]{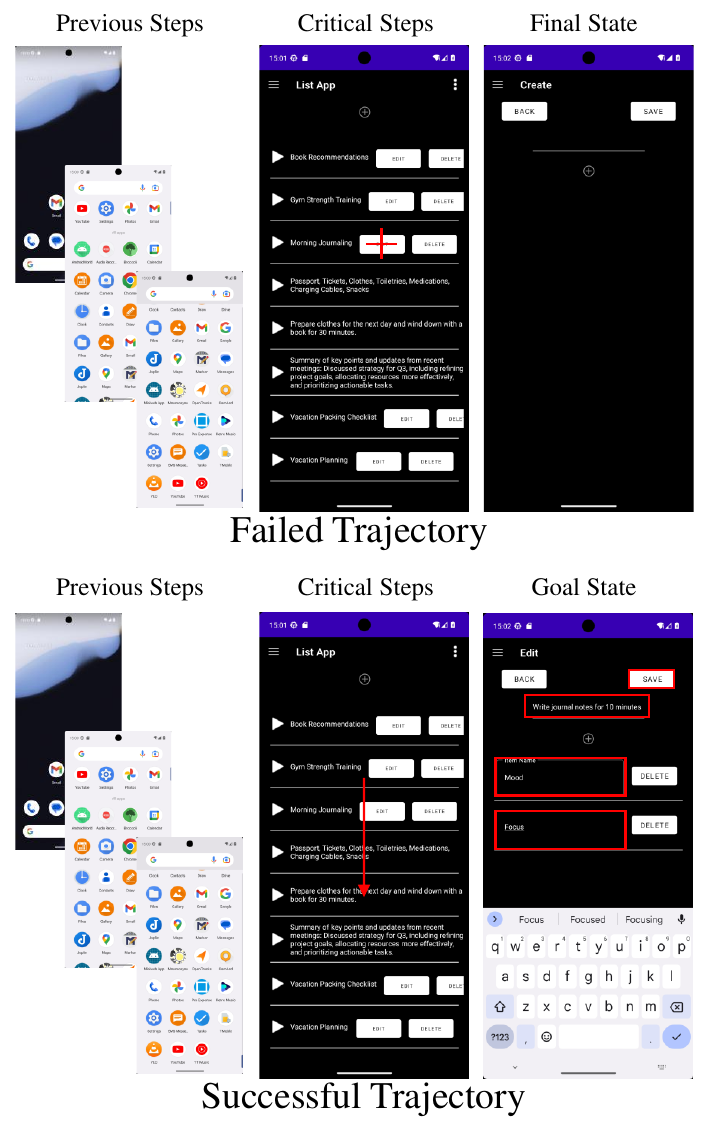}
\caption{Comparison of UI-TARS-trained agents for task: In Mnemosyne, open the "Write journal notes for 10 minutes" item, set the first entry to "Mood" and the second entry to "Focus", without saving it.}
\label{fig:case_online_1}
\end{figure}

\section{Case Study}
\label{appendix:case_study}

\subsection{Case Study for Self-Evolving}

We show examples of the self-evolving mechanism illustrating how it modifies the environment state during the iterative process in Figure~\ref{fig:case_self_evolve}. It can be clearly observed that the app's main pages become increasingly diverse during the later stages of evolution, enabling the collection of more complex and varied tasks for subsequent training.

\subsection{Case Study for Online Performance}

\begin{figure}[t!]
\centering
\includegraphics[width=\linewidth]{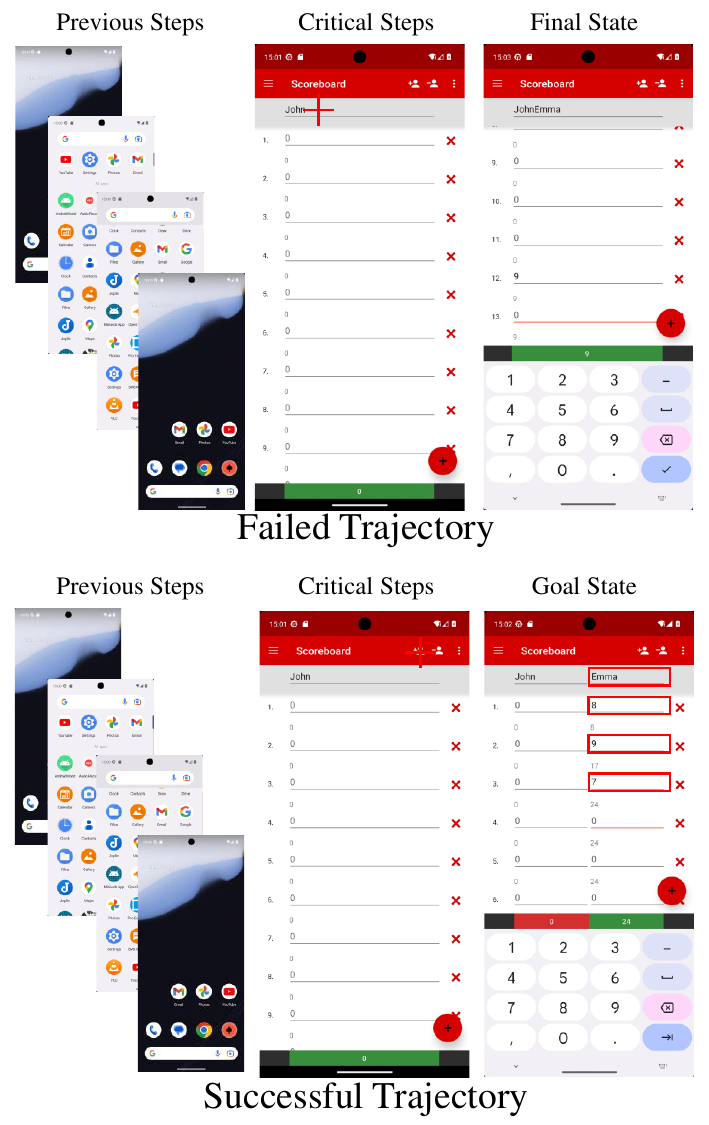}
\caption{Comparison of GUI-Owl-trained agents for task: Add a new player named Emma to the Scoreboard app, and set her scores to 8 points in Round 1, 9 points in Round 2, and 7 points in Round 3.}
\label{fig:case_online_2}
\end{figure}

To further illustrate the impact of different reward methods on agent behavior during online interactions, we present two representative case studies in Figure~\ref{fig:case_online_1} and Figure~\ref{fig:case_online_2}. One shows a failed trajectory from an agent trained with StepCritic, while the other demonstrates a successful trajectory from an agent trained with~\name. These examples highlight how using our reward framework enables agents to perform complex mobile GUI tasks more reliably and accurately.

\section{Prompts}
\label{appendix:prompts}

\subsection{Prompting Template of Task Generation}
\begin{prompt}
You are an expert in analyzing mobile app user interfaces and generating clear, purposeful, and executable task instructions.
You will receive:
- The app name
- A current mobile screenshot
- UI component information extracted from the screenshot
- The action history from app launch to the current screen

Your goal is to analyze the current user interface and generate purposeful instructions based on the action history, representing the tasks a human user is likely to perform next.
\end{prompt}
\begin{prompt}
Task types:
- Task-Oriented: A series of operations to achieve a specific goal.
- Question-Oriented: A series of operations that lead to answering a specific question.

Requirements:
- Each task must mention the app name and provide a high-level instruction without referencing any specific UI elements.
- You can combine multiple simple tasks (from the example tasks) into richer, more complex, and more diverse tasks.
- If the task involves entering information, specify the exact content to be input.
- DO NOT use words such as 'or', 'any', 'e.g.', or any other expression that makes the task description vague or ambiguous.
- DO NOT mention any non-existent files.
- DO NOT generate tasks related to user login, account registration, account creation, or authentication flows.
- Do not generate any tasks if the screen is a permission dialog, an app initialization popup, or unrelated to the current app.
- Generate between 0 and \{\texttt{max\_tasks}\} tasks depending on the complexity of the screen and action history.

Context:
- App name: \{\texttt{app\_name}\}
- UI components: \{\texttt{accessibility\_tree}\}
- Action history: \{\texttt{action\_summary}\}

Example tasks: \{\texttt{examples}\

Output format (strictly follow this format):
your analysis:
- A brief explanation of the navigation history, user intent, and key UI elements that affect the tasks.
your tasks:
1. Task description
2. Task description
3. Task description
...
Do not add any extra sections or commentary outside this format.
\end{prompt}

\subsection{Prompting Template of Task Inherit}
\begin{prompt}
You are an expert at generating follow-up mobile app tasks based on an existing completed task.

You will receive:
- A user task trajectory, including the initial user goal, the last few screenshots, and a sequence of action summaries that indicate the task has already been completed.

Your goal is to generate a new task that STARTS FROM the environment state after the previous task is finished.
The new task should build upon the existing result and introduce additional steps or subtasks to continue the workflow.

Requirements:
- The new task must assume the previous task has already been successfully completed.
- The task must start from the current environment state shown in the screenshots.
- The task should extend the workflow by adding new steps or subtasks, without repeating the original task.
- The task must mention the app name and provide a high-level instruction without referencing any specific UI elements.
- If the task involves entering information, specify the exact content to be input.
- DO NOT use words such as 'or', 'any', 'e.g.', or any other expression that makes the task description vague or ambiguous.
- DO NOT mention any non-existent files.

Input:
- App name: \texttt{\{app\_name\}}
- Previous completed user goal: \texttt{\{initial\_goal\}}
- Action summary: \texttt{\{action\_summary\}}
- Current environment screenshots: <image>

Output format (strictly follow this format):
Your analysis:
- A brief explanation of how the new task continues from the completed state and what new steps are added.
Output task:
<task>
(contain only a single task; if no task needs to be generated, write "No task")
</task>

Do not add any extra sections or commentary outside this format.

\end{prompt}

\subsection{Prompting Template of Task Merge}
\begin{prompt}
You are an expert at generating more complex mobile app tasks by merging an existing task with additional subtasks.

You will receive:
- A user task trajectory, including the initial user goal, the last few screenshots, and a sequence of action summaries.

Your goal is to generate a NEW combined task that integrates the original task and newly added steps into a single, more complex task.
The initial environment of the new task corresponds to the environment state shown in the screenshots.

Requirements:
- The new task must explicitly include the original task goal and extend it with additional steps or subtasks.
- The task should describe a complete, merged workflow as a single high-level user goal.
- The task must mention the app name and provide a high-level instruction without referencing any specific UI elements.
- If the task involves entering information, specify the exact content to be input.
- DO NOT use words such as 'or', 'any', 'e.g.', or any other expression that makes the task description vague or ambiguous.
- DO NOT mention any non-existent files.

Input:
- App name: \texttt{\{app\_name\}}  
- Original user goal: \texttt{\{initial\_goal\}}  
- Action summary: \texttt{\{action\_summary\}}
- Initial environment screenshots: <image>

Output format (strictly follow this format):
Your analysis:
- A brief explanation of how the original task and new subtasks are merged into a more complex task.
Output task:
<task>
(contain only a single task; if no task needs to be generated, write "No task")
</task>

Do not add any extra sections or commentary outside this format.
\end{prompt}

\subsection{Prompting Template of Task Change}
\begin{prompt}
You are an expert at generating variant mobile app tasks by modifying parameters in an existing task instruction.

You will receive:
- A user task trajectory, including the initial user goal, the last few screenshots, and a sequence of action summaries.

Your goal is to generate a new task that has the SAME task structure as the original task,
but differs by changing specific parameters such as names, values, text content, or identifiers.

Requirements:
- The new task must keep the same overall workflow and steps as the original task.
- Only modify concrete parameters in the task instruction, such as file names, text content, titles, numbers, or labels.
- Do NOT add new steps or remove existing steps.
- The task must mention the app name and provide a high-level instruction without referencing any specific UI elements.
- If the task involves entering information, specify the exact content to be input.
- DO NOT use words such as 'or', 'any', 'e.g.', or any other expression that makes the task description vague or ambiguous.
- DO NOT mention any non-existent files.

Input:
- App name: \texttt{\{app\_name\}}  
- Original user goal: \texttt{\{initial\_goal\}}  
- Action summary: \texttt{\{action\_summary\}}
- Environment screenshots: <image>

Output format (strictly follow this format):

Your analysis:
- A brief explanation of which parameters are modified and how they differ from the original task.
Output task:
<task>
(contain only a single task; if no task needs to be generated, write "No task")
</task>

Do not add any extra sections or commentary outside this format.
\end{prompt}

\subsection{Prompting Template of label anchor}
\begin{prompt}
You are an expert in GUI understanding. Your task is to analyze the input image and UI Elements, and select the indices of the UI elements that are relevant to the instruction. These selected UI elements should uniquely identify the state after the instruction task has been completed.

Input Description:
- Instruction: A natural language description of the GUI task to be performed.
- UI Elements: The complete list of UI elements extracted from the final-state screen (including all element metadata).
- Image: The screenshot of the final state after completing the task, with UI element bounding boxes and indices overlaid to indicate their positions.

Requirements:
- Compare the image and the UI Elements list, and find UI elements whose content or visual role reflects the completion of the instruction.
- Select key UI elements that can indicate the final page state, such as navigation bars, menu titles, tab titles, selected buttons, status indicators, etc.
- Avoid selecting UI elements whose state may change dynamically due to time or network content.
- If you believe that no UI element in the screen is relevant to the instruction, stop the reasoning and output "No related elements".

Now output in the following format:
Thinking Process: Your thinking process
Output Index: [x, x, ...]

Input:
Instruction: \texttt{\{query\}}  
UI Elements: \texttt{\{ui\_elements\_info\}}
\end{prompt}

\end{document}